\documentclass[conference]{IEEEtran}
\IEEEoverridecommandlockouts
\usepackage{cite}
\usepackage{amsmath,amssymb,amsfonts}
\usepackage{graphicx}
\usepackage{textcomp}
\usepackage{xcolor}
\usepackage{array}
\usepackage{makecell}
\usepackage{algorithm} 
\usepackage{hyperref} 
\usepackage[noend]{algpseudocode}
\usepackage{fancyhdr}

\usepackage{blindtext}

\DeclareMathOperator{\DPQ}{DPQ}

\def\BibTeX{{\rm B\kern-.05em{\sc i\kern-.025em b}\kern-.08em
    T\kern-.1667em\lower.7ex\hbox{E}\kern-.125emX}}
\begin{document}

\title{Permutation Learning with Only N Parameters: From SoftSort to Self-Organizing Gaussians
}

\author{\IEEEauthorblockN{Kai Uwe Barthel}
\IEEEauthorblockA{\textit{Visual Computing Group} \\
\textit{HTW Berlin}\\
Berlin, Germany \\
0000-0001-6309-572X}
\and
\IEEEauthorblockN{Florian Tim Barthel}
\IEEEauthorblockA{\textit{Vision and Imaging Technologies} \\
\textit{Fraunhofer HHI}\\
Berlin, Germany \\
0009-0004-7264-1672}
\and
\IEEEauthorblockN{Peter Eisert}
\IEEEauthorblockA{\textit{Vision and Imaging Technologies} \\
\textit{Fraunhofer HHI}\\
Berlin, Germany \\
0000-0001-8378-4805}
}

\maketitle

\begin{abstract}
Permutation learning is essential for organizing high-dimensional data in optimization and machine learning. Current methods like Gumbel-Sinkhorn require $\mathbf{N^2}$ parameters for $\mathbf{N}$ objects, operating on the full permutation matrix. While low-rank approximations offer some reduction to $\mathbf{2NM}$ (with $\mathbf{M \ll N}$), they still create a computational bottleneck for very large datasets. SoftSort, a continuous relaxation of the argsort operator, enables differentiable 1D sorting but struggles with multidimensional data and complex permutations. We introduce a novel method for learning permutations using only $\mathbf{N}$ parameters, dramatically reducing storage costs. Our method extends SoftSort by iteratively shuffling the $\mathbf{N}$ indices of the elements to be sorted and applying a few SoftSort optimization steps per iteration. This significantly improves sorting quality, especially for multidimensional data and complex criteria, outperforming pure SoftSort. Our method offers superior memory efficiency and scalability while maintaining high-quality permutation learning. Its drastically reduced memory requirements make it ideal for large-scale optimization tasks like "Self-Organizing Gaussians", where efficient and scalable permutation learning is critical.
\end{abstract}


\section{Introduction \& Related Work}
Learning permutations is a fundamental challenge in machine learning, computer vision, and optimization. Many real-world problems, such as ranking, assignment, and sorting, require finding an optimal arrangement of elements \cite{sinkhorn1964relationship, santa2018visual, petersen2022monotonic}. The challenge in efficiently learning permutations lies in their discrete, binary matrix representation (one '1' per row/column), which prevents direct gradient-based optimization.
 
\begin{figure}[t]
\centerline{\includegraphics[width=1\linewidth]{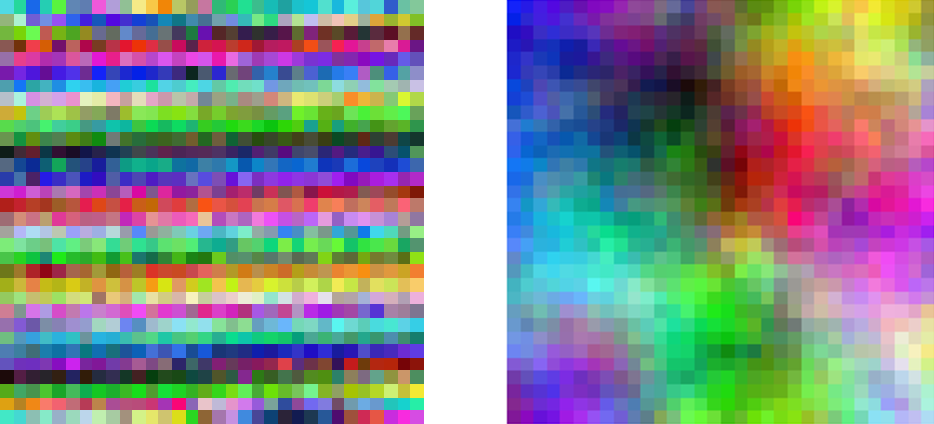}}
\caption{\label{fig:color_example} Example of grid-based sorting for 1024 random RGB colors sorted by \textit{SoftSort} (left) and the new proposed approach using  \textit{ShuffleSoftSort} (right). The loss function minimizes the average color distance of neighboring grid cells.}
\end{figure}

While our proposed permutation learning method has broad applicability across various fields, this paper focuses on its use in distance-preserving grid layout algorithms for color or visual image sorting (Figures \ref{fig:color_example} / \ref{fig:kitchen}) and 3D scene reconstruction in 3D Gaussian Splatting \cite{kkl23} (Figure \ref{fig:sog}). Sorting images based on similarity enhances visual perception, enabling users to efficiently explore hundreds of images simultaneously. In 3D scene reconstruction, organizing each scene attribute into a sorted 2D grid increases spatial correlation, and the resulting high-dimensional maps can be compressed using standard codecs, achieving substantial reductions without sacrificing rendering quality.

\subsection{Permutation Learning}\label{PL}
\begin{figure}[h]
\centerline{\includegraphics[width=1\linewidth]{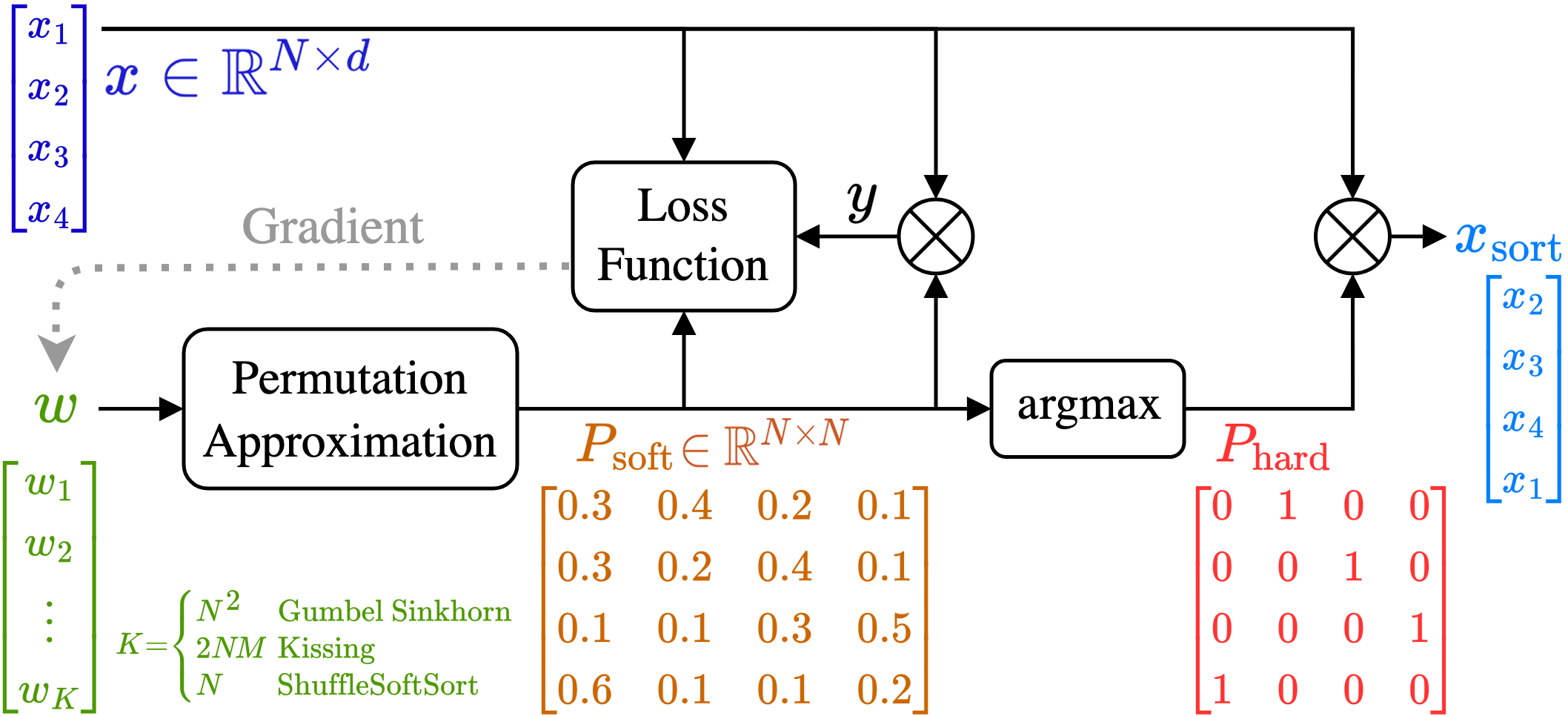}}
\caption{\label{fig:method_overview}
Permutation learning optimizes the differentiable permutation matrix \( P_{\text{soft}} \) by adjusting the weights \( w \) based on a given loss function. The number of weights, $K$, depends on the chosen permutation approximation method. The input vectors \( x \) are then reordered into \( x_{\text{sort}} \) using \( P_{\text{hard}} \), which is obtained by applying a row-wise argmax operator to \( P_{\text{soft}} \).
}
\end{figure}

Permutation learning aims to estimate a permutation matrix $P_{\text{hard}}$ that sorts input vectors $x$ into a desired order $x_{\text{sort}}$ according to a predefined criterion. As depicted in Figure \ref{fig:method_overview}, the fundamental challenge in this process is that matrix multiplication with the discrete $P_{\text{hard}}$ is not differentiable, which prevents direct optimization. To circumvent this, a continuous relaxation, $P_{\text{soft}}$, is optimized via a carefully designed loss function. This function's purpose is twofold: to ensure that the final binarized $P_{\text{soft}}$ (yielding $P_{\text{hard}}$) is a valid permutation and to guarantee it meets the specified optimization criteria.

This following discusses various differentiable permutation learning approaches, from existing methods to our proposed \textit{ShuffleSoftSort}.

One of the most well-established approaches is the \textit{Gumbel-Sinkhorn} method \cite{mena2018learning}. This technique relaxes discrete permutations into doubly stochastic matrices and applies Sinkhorn normalization, allowing for end-to-end differentiable learning of approximate permutations. While Gumbel-Sinkhorn can generate high-quality permutations, it requires storing and computing an \(N^2\) matrix, where $N$ is the number of elements to be sorted. This quadratic memory consumption makes it impractical for large-scale problems.

To mitigate these high memory requirements, \textit{low-rank factorization techniques} have been proposed, such as the "Kissing to Find a Match" method by Droge et al. \cite{kissing}. This approach approximates permutation matrices using a low-rank decomposition, typically involving two row-normalized matrices $V$ and $W$ of size $N \times M$, where $M \ll N$. The permutation matrix is then approximated by $P \approx V W^T$, followed by scaling and row-wise softmax normalization. These methods require only $2N\!M$ parameters, significantly reducing memory usage compared to Gumbel-Sinkhorn, though they may still be too memory-intensive for very large problems.

An alternative is \textit{SoftSort} \cite{softsort}, a continuous relaxation of the \textit{argsort} operator, which enables differentiable sorting. Unlike Gumbel-Sinkhorn or low-rank methods, SoftSort requires only $N$ parameters, making it highly memory-efficient. However, its inherent limitation to 1D sorting restricts its use in complex permutation tasks like assignment problems or distance-preserving grid layouts.

In this paper, we extend SoftSort to learn permutations with only $N$ parameters, specifically addressing its quality limitations and enabling multi-dimensional sorting. Our proposed \textit{ShuffleSoftSort} technique overcomes these challenges.

\subsection{Heuristic Distance-Preserving Grid Layout Algorithms}

Distance-Preserving Grid Layouts optimize the assignment of objects or images to grid positions by aligning spatial proximity with the similarity relationships of their feature vectors.  
Various algorithms have been proposed for arranging vectors on a 2D grid, most of which rely on heuristic methods that offer efficient performance. However, recent research \cite{GradSort} has demonstrated that gradient-based learning techniques can produce sorted grid layouts with superior sorting quality compared to traditional approaches.  

In the following sections, we present a brief overview of the most commonly used non-learning methods.

A \textit{Self-Organizing Map} (SOM) \cite{Kohonen1982, Kohonen2013} uses a grid of map vectors with the same dimensionality as the input vectors. It assigns these vectors to the most similar grid position and iteratively updates the map vectors based on their neighborhood.

A \textit{Self-Sorting Map} (SSM) \cite{Strong2011, Strong2014} initializes grid cells with input vectors and employs a hierarchical swapping process. This method compares vector similarity with the average of their grid neighborhood, considering all possible swaps within a set of four cells.

\textit{Linear Assignment Sorting} (LAS) \cite{Improved_Wiley} merges the concepts of SOM (using a continuously filtered map) and SSM (performing cell swaps), extending the idea to optimally swap all vectors simultaneously. \textit{Fast Linear Assignment Sorting} (FLAS) improves runtime efficiency by iterative swapping subsets, achieving sorting quality close to LAS.

\textit{Dimensionality reduction methods} like t-SNE \cite{Maaten2008} and UMAP \cite{McInnes2018} can project high-dimensional vectors onto a 2D plane before grid arrangement. Linear assignment solvers, such as the Jonker–Volgenant algorithm \cite{Jonker1987}, then map these positions to optimal grid locations. Fast placement strategies are discussed in \cite{Hilasaca2021}.

\subsection{Contributions}
In \cite{GradSort}, we introduced the first gradient-based grid sorting method using Gumbel-Sinkhorn. It ensures a valid permutation matrix while optimizing the grid arrangement for vector similarity, achieving superior sorting quality. However, its \(\mathcal{O}(N^2)\) memory demand limits scalability for large datasets.
To address this, we introduce \textit{ShuffleSoftSort}, a novel SoftSort extension for high-performance, memory-efficient permutation learning. 
Our key contributions are:

\begin{itemize}
    \item \textbf{Memory Efficiency:} ShuffleSoftSort uses only $N$ parameters. It enhances SoftSort via iterated shuffling of indices, enabling large-scale swaps and improving sorting quality.
    \item \textbf{Superior to Low-Rank Methods:} ShuffleSoftSort outperforms low-rank approximations like \textit{Kissing to Find a Match} with significantly lower memory consumption.
    \item \textbf{Scalable for Large-Scale Applications:} Drastically reducing memory needs compared to Gumbel-Sinkhorn while maintaining strong performance, ShuffleSoftSort enables scalable permutation learning for large optimization tasks, e.g., \textit{Self-organizing Gaussians} and data visualization.
\end{itemize}



\section{Proposed Method}
Our new approach is based on \textit{SoftSort}, which was originally developed as a continuous relaxation for the \textit{argsort} operator, defined as 
\begin{equation}
    SoftSort_\tau^\mathrm{D}(w)=softmax\left(\frac{-\mathrm{D}(w_{\mathrm{sort}},w)}{\tau}\right),
\end{equation}
with $\mathrm{D}(w_{\mathrm{sort}},w) $ being the L1 distance matrix between the sorted and the unsorted elements of the vector $w$. Row-wise softmax increases matrix elements at positions where elements coincide, while suppressing the others. With decreasing $\tau$ the matrix converges to the permutation matrix.
SoftSort produces valid permutations without iterated normalization, but is poorly suited for high-dimensional data due to its inherent one-dimensional sorting. 

\begin{figure}[t]
\centerline{\includegraphics[width=1\linewidth]{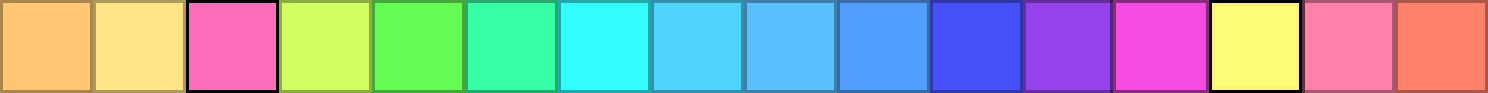}}
\caption{\label{fig:1D_color_toy_example} 
1D color arrangement highlighting SoftSort's challenges (see text).
}
\end{figure}

A limitation of SoftSort is illustrated by the 1D color toy example in Figure \ref{fig:1D_color_toy_example}. Achieving better color ordering (by swapping yellow and magenta) necessitates traversing intermediate positions that initially degrade quality, causing gradient-based optimization to fail. 
Figure \ref{fig:color_example} (left) further demonstrates this in a 2D grid-based RGB color sorting scenario. Because SoftSort uses a one-dimensional weight vector to represent element order, any learned position change is restricted to this single dimension, proving inadequate for complex reordering across several rows.

Our newly proposed approach, \textit{ShuffleSoftSort}, retains the advantages of SoftSort, such as low memory usage and the ability to easily achieve valid permutations, while improving the quality of the permutations compared to SoftSort. 
Since SoftSort can only perform one-dimensional sorting, it struggles with problems like those shown in Figures \ref{fig:color_example} or \ref{fig:1D_color_toy_example}. 
To solve this problem, we circumvent the one-dimensional constraint without altering the fundamental principle.

Our method iteratively reorganizes the elements' indices and then applies SoftSort to these updated one-dimensional indices. By dynamically reordering the 1D input, we enable element repositioning not achievable with a static input order.

\begin{figure}[h!]
\centerline{\includegraphics[width=1\linewidth]{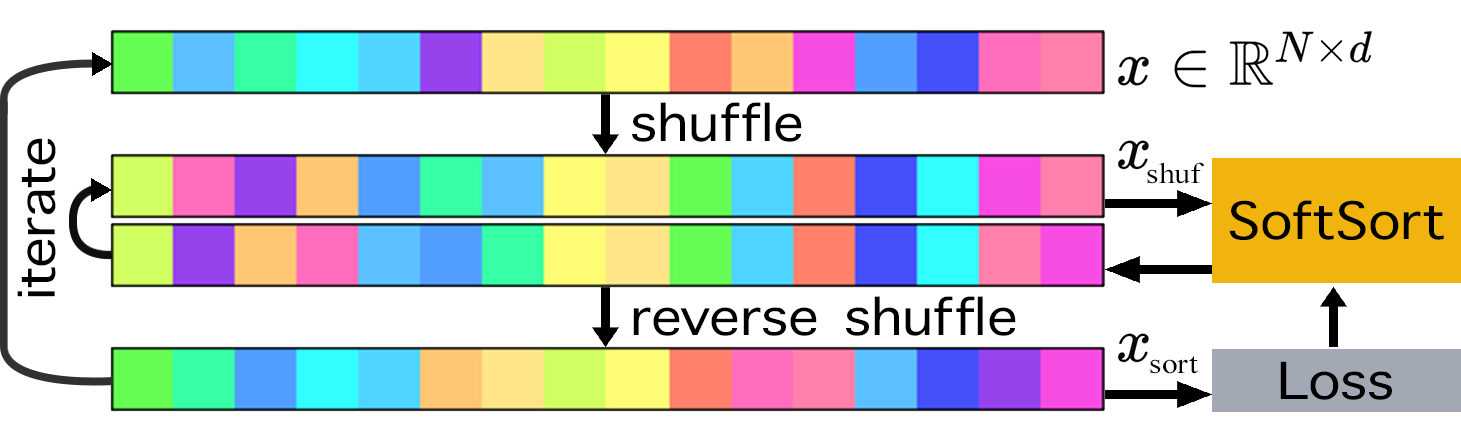}}
\caption{\label{fig:indices_swap} 
ShuffleSoftSort enhances sorting by iteratively applying SoftSort to randomly shuffled elements. The loss is computed on the reverse-shuffled output, refining the permutation and addressing SoftSort's limitations.
\\
}
\end{figure}

Figure~\ref{fig:indices_swap} highlights ShuffleSoftSort's core concept, demonstrating how its repeated shuffled index reorganization dramatically improves sorting flexibility. Notably, the loss function is computed on reverse-shuffled elements. Training involves gradually decreasing $\tau$, which minimizes the influence of more distant elements. Consequently, sorting quality improves with an increased number of iterations ($R$). Algorithm~\ref{alg:SSS} presents the ShuffleSoftSort procedure, which sorts the $N$ elements of a vector $x$, where each element has dimension $d$.

\begin{algorithm} [h!] \footnotesize 
\caption{ShuffleSoftSort}\label{alg:SSS}
\begin{algorithmic}[0]  
\vspace{0.2em}
    \For{$r = 0$ to $R$} \Comment{Perform global iterations}
        \vspace{0.2em}
        \State $\tau = \tau_{\text{start}} ( \frac{\tau_{\text{end}}}{\tau_{\text{start}}} )^{\frac{r}{R}}$ \Comment{Decrease $\tau$ from $\tau_{\text{start}} \text{ to } \tau_{\text{end}}$}
        \vspace{0.2em}
        \State $w$ = arange(0, N) \Comment{Initialize SoftSort weights $w$ linearly}
        \vspace{0.2em}
        \State \textcolor{blue}{shuf\_idx} = randperm(N) \Comment{Generate random shuffle indices}
        \vspace{0.2em}
        \State x\_shuf = x[\textcolor{blue}{shuf\_idx}] \Comment{Apply shuffle to input vector $x$} 
        \vspace{0.4em}
        \For{$i = 1$ to $I$} \Comment{ Perform  SoftSort inner optimization steps}
            \vspace{0.2em}
            \State perm\_soft = soft\_sort($w$, $\tau$) \Comment{Compute soft permutation matrix}
            \vspace{0.2em}
            \State x\_sort\_soft = perm\_soft @ x\_shuf \Comment{Apply soft permutation}
            \vspace{0.2em}
            \State x\_sort\_soft[\textcolor{blue}{shuf\_idx}]  = x\_sort\_soft \Comment{Reverse shuffle for loss}
            \vspace{0.2em}
            \State get\_loss(x\_sort\_soft, perm\_soft).backward() \Comment{Loss backward pass}
        \EndFor
        \vspace{0.2em}
        \State \textcolor{violet}{sort\_idx} = $argmax$(perm\_soft, -1) \Comment{Extract hard permutation indices}
        \vspace{0.2em}
        \State x\_sort = x[\textcolor{blue}{shuf\_idx}]  = x\_shuf[\textcolor{violet}{sort\_idx}] \Comment{Apply hard permutation}
    \EndFor

\end{algorithmic}
\end{algorithm}


For grid-based sorting with ShuffleSoftSort, we use a loss function that can be efficiently computed in a separable manner.
It builds on \cite{GradSort}, incorporating the \textit{neighborhood loss} \( L_{nbr} \) (the normalized average distance of neighboring grid vectors in horizontal and vertical directions). However, it avoids the computationally and memory-intensive distance matrix loss. We define our loss function as follows (using $P$ for $P_{\text{soft}}$):
\begin{equation} \label{eq:loss2}
L(P) = \underbrace{L_{nbr}(P)}_{\text{smoothness term}} + \;\;\; \underbrace{\lambda_s L_s(P) + 
\lambda_{\sigma} L_{\sigma}(P) }_{\text{regularization terms}} \ .
\end{equation}

We obtained the best results by combining two loss components to generate valid permutation matrices. 
The first regularization term, the \textit{stochastic constraint loss} $L_s(P)$, ensures that \( P_{\text{soft}} \) converges to a doubly stochastic matrix by penalizing deviations from 1 in the column sums of the matrix:
\begin{equation}
L_s(P)=\frac{1}{N}\sum_j \Bigl( \Bigl(\sum_i P_{ij}\Bigr)-1  \Bigr)^2  
\end{equation}

The $L_{\sigma}(P)$ term is the \textit{standard deviation loss}. It encourages $P_{\text{soft}}$ to preserve original feature statistics by minimizing the sum of absolute differences between the column-wise standard deviations of $x$ and $y = P_{\text{soft}} \cdot x$, normalized by the sum of $\boldsymbol{\sigma}_x$.
$\boldsymbol{\sigma}_x$ and $\boldsymbol{\sigma}_y$ are the vectors of column-wise standard deviations for $x$ and $y$, respectively.

\begin{equation}  L_{\sigma}(P) = \frac{\sum | \boldsymbol{\sigma}_x - \boldsymbol{\sigma}_y |}{\sum \boldsymbol{\sigma}_x} 
\end{equation}

The following table summarizes the properties of the discussed permutation approximation methods, along with the characteristics of our proposed \textit{ShuffleSoftSort} scheme. Stability here means how certain it is to obtain a valid permutation matrix without duplicates.

\begin{table}[h]
\setlength{\extrarowheight}{1.5pt}
\centering
\begin{tabular}{|l|c|c|c|c|}
\hline
 & \makecell{Gumbel-\\Sinkhorn} & Kissing & SoftSort & \textbf{Ours} \\
\hline
Number of Parameters K & $N^2$ & $2N\!M$ & $N$ & $N$ \\
\hline
Non-iterative normalization & no & yes & yes & yes \\
\hline
Quality & ++ & + & -- & ++ \\
\hline
Stability & + & o & ++ & ++ \\
\hline
\end{tabular}

\vspace{0.2cm}
\label{tab:methods}
\caption{Comparison of permutation approximation methods}
\end{table}

The proposed new sorting strategy extends to other permutation learning tasks, with the key idea of continuously modifying the index order to allow for greater flexibility in sorting.
For a memory-efficient implementation, it is crucial to compute the permutation matrix and the loss components in a row-wise manner, as the complete storage would require \(N^2\) elements.

\section{Experimental Evaluation}


In this section, we compare Gumbel-Sinkhorn, Kissing to Find a Match, and SoftSort with our newly introduced ShuffleSoftSort. The evaluation is conducted using 1024 randomly generated RGB colors. For consistency, the loss function from \cite{GradSort} is employed for the first three methods, while ShuffleSoftSort uses its adapted version from Eq.~\ref{eq:loss2}.

Training parameters were set as follows: regularization parameters $\lambda_s = 1$ and $\lambda_{\sigma} = 2$; a learning rate of 0.3; $\tau_{\text{start}} = 1$ and $\tau_{\text{end}} = 0.1$. The global iterations $R$ were set to 2500, and the inner SoftSort iterations $I$ to 3. While ShuffleSoftSort's inner SoftSort loop typically yields a valid permutation matrix, SoftSort iterations are extended in very rare cases where permutation matrix columns contain duplicates, ensuring a valid permutation is ultimately achieved.

In grid-based sorting scenarios, ShuffleSoftSort's performance can be further enhanced. The shuffling process can be adapted to progressively reduce the range of possible positional changes during training. This ensures that towards the end of training, only closely located positions are swapped. Furthermore, it's beneficial to alternate the indexing of 2D positions between row-wise and column-wise. This strategy enables fine-grained positional adjustments in both horizontal and vertical directions.

The table below compares these methods based on memory requirements for learnable parameters, runtime on an Apple M1 Max, and sorting quality, measured by Distance Preservation Quality (\(\DPQ_{16}\)). 
As demonstrated in [3], \(\DPQ_{16}\) is a perceptually driven metric that closely aligns with human visual judgment and strongly correlates with the mean similarity to neighboring elements.
Figure \ref{fig:color_example} shows the images obtained with SoftSort and ShuffleSoftSort.

\begin{table}[h]
\setlength{\extrarowheight}{1.5pt}
\centering
\begin{tabular}{|p{0.33\linewidth}|c|c|c|}
\hline
\textbf{Method} & \textbf{Memory $\downarrow$} & \textbf{Runtime [s] $\downarrow$} & \textbf{Quality $\uparrow$} \\
\hline
Gumbel-Sinkhorn \cite{mena2018learning} & 1048576 & 226.8 & \textbf{0.913}  \\
\hline
Kissing \cite{kissing} & 26624 & 114.4 & $-^*$  \\
\hline
SoftSort \cite{softsort} & \textbf{1024} & 110.7 & 0.698 \\
\hline
\textbf{ShuffleSoftSort (ours)} & \textbf{1024} & \textbf{91.5} & 0.909  \\
\hline
\multicolumn{4}{r}{*) invalid permutation}
\end{tabular}
\vspace{0.2cm}
\label{tab:evaluation}
\caption{Evaluating Permutation Learning Methods on Color Sorting}
\end{table}

Further experiments are needed to optimize the parameter settings. However, this table already demonstrates the overall feasibility and effectiveness of the proposed approach, achieving high quality with lowest memory and computational demands.
The runtimes are not optimal and only have a relative meaning, as the implementations were not created as optimized GPU applications. ShuffleSoftSort benefits from the fact that its loss function is considerably simpler than that of the other schemes.  
Notably, the kissing approach exhibits poor convergence due to its simple softmax normalization, often failing to produce valid permutation matrices.


\section{Applications}

\subsection{Grid-based Image Sorting}
Viewing large volumes of images is a cognitive challenge, as human perception becomes overwhelmed when too many images are displayed at once. To address this, most applications limit the visible images to around 20. However, sorting images based on the similarity of their visual feature vectors enhances navigation, allowing users to view hundreds of images simultaneously. This method is particularly useful for stock agencies and e-commerce platforms, where efficient browsing is key.

\begin{figure}[htb]
\centerline{\includegraphics[width=1.0\linewidth]{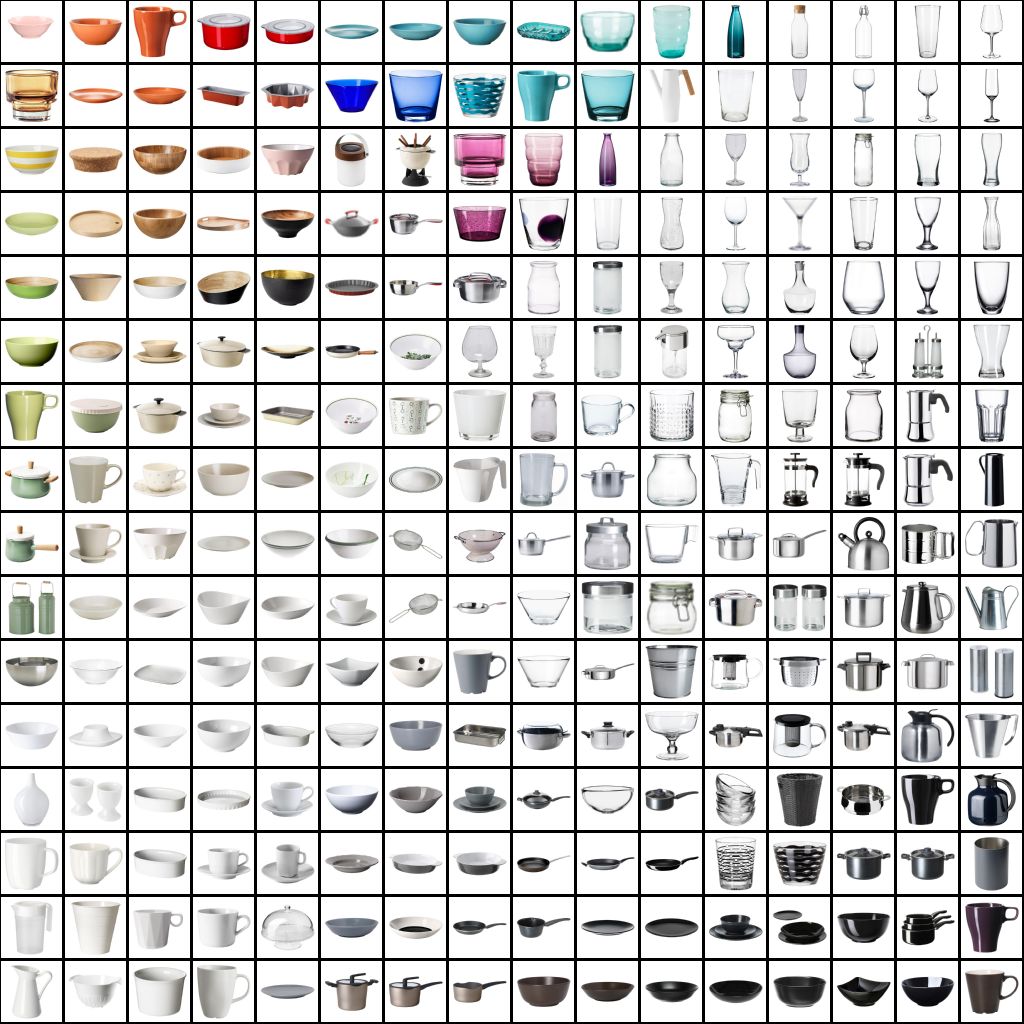}}
\caption{Sorting example of a dataset of e-commerce images, simplifying navigation, browsing, and retrieval of large image databases.
}
\label{fig:kitchen}
\end{figure}

Visual feature vectors, generated through image analysis or deep learning, help organize images by content. Low-level feature vectors, with tens of dimensions, are often more effective for sorting larger image sets, as they enable users to group similar images easily. In that study \cite{Improved_Wiley} and the sorting shown in Figure \ref{fig:kitchen}, we used 50-dimensional low-level feature vectors to capture the key visual properties for sorting.
Applying grid-based sorting to these vectors results in organized, visually meaningful layouts, as shown below.

\subsection{Self-Organizing Gaussians}
\begin{figure}[htb]
\centerline{\includegraphics[width=0.9\linewidth]{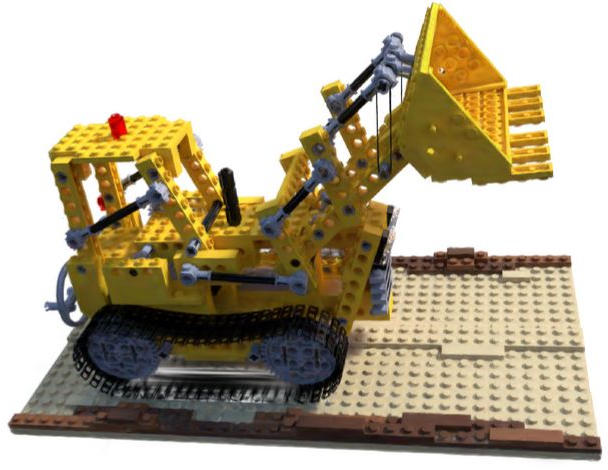}}
\centerline{\includegraphics[width=0.9\linewidth]
{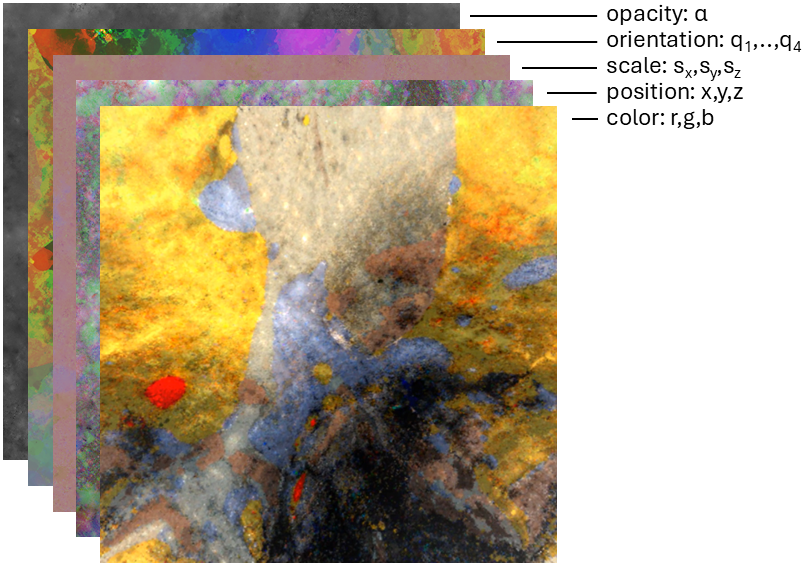}}
\caption{
Top: photorealistic rendering of a Gaussian Splatting scene; bottom: same Gaussians with all attributes sorted into a 2D grid using \cite{morgenstern2024compact}, which enforces high spatial correlation of points for efficient compression.
}
\label{fig:sog}
\end{figure}

Another application for large scale permutation learning can be found in the field of 3D scene reconstruction, specifically in \textit{3D Gaussian Splatting} (3DGS) \cite{kkl23}. 3DGS has found a lot of attention in recent years, as it solves several challenges in 3D scene reconstruction, such as rendering performance, photorealism, explicitness, or explainability compared to prior methods. On the downside, however, 3DGS suffers from very large storage sizes. This is because 3DGS stores a 3D scene as an unstructured point cloud with millions of data points, each consisting of several parameters like position, scale, orientation, opacity, base color and optionally spherical harmonics. As a result, larger scenes can take up several gigabytes of storage, making them less portable. This has motivated significant research  \cite{bkl25} for efficient compression of these datasets.
In contrast to other multimedia data, 3DGS data exhibits an interesting property, namely the ambiguity for point ordering. Any reshuffling of Gaussian Splats will lead to exactly the same results, which has been exploited by \textit{Self-Organizing Gaussians} (SOG) \cite{morgenstern2024compact}. Instead of storing the data points in vectors, SOG creates a sorted 2D grid for each of the scenes attributes, as shown in Figure \ref{fig:sog}, increasing spatial correlation of neighboring points. By incorporating a 2D smoothness loss also in the optimization process of the ambiguous splat configurations, further gains can be achieved.
The high dimensional maps are then compressed using standard image compression codecs, allowing for up to 40x storage reduction without compromising in rendering quality. The original SOG uses a heuristic non-differentiable sorting, given the large number of data points $N$. However, the proposed method offers gradient-based permutation learning also for such large datasets, requiring only the storage of $N$ parameters instead of $N^2$. This allows for optimally sorting millions of data points without exceeding the memory capacity and enabling end-to-end learning for scene reconstruction.

\section{Conclusion}
This work presented ShuffleSoftSort, a novel approach to permutation learning designed for high sorting quality with drastically reduced memory requirements. Our method achieves this by employing an iterative shuffling mechanism, enabling efficient $N$-parameter permutation learning, a significant advance over Gumbel-Sinkhorn and low-rank approximations. The demonstrated strong performance and superior memory efficiency make ShuffleSoftSort highly valuable for diverse large-scale optimization tasks, including Self-Organizing Gaussians, data visualization, and other complex permutation problems. Its versatility is further exemplified by applications like solving Sudokus, which can be explored in our GitHub repository. Future research will focus on further optimizing and extending these differentiable sorting techniques for multidimensional applications.
Our code and experiments can be found at:

\href{https://github.com/Visual-Computing/ShuffleSoftSort}{https://github.com/Visual-Computing/ShuffleSoftSort}

\bibliographystyle{abbrv-doi}
\bibliography{conference_101719}

\end{document}